\documentclass[11pt]{article}
\usepackage{konvens2016}
\usepackage{mathptmx}
\usepackage[scaled=.90]{helvet}
\usepackage{courier}
\usepackage{url}
\usepackage{latexsym}
\usepackage{xcolor}
\usepackage{latexsym}
\usepackage{tabularx}
\usepackage[german, english]{babel}
\usepackage{multirow}
\usepackage{booktabs}
\usepackage{ifthen}
\usepackage{tikz,pgfplots}
\pgfplotsset{compat=newest}
\usepackage{xcolor,colortbl}
\usepackage[color=cyan]{todonotes}
\usepackage{latexsym}
\usepackage{graphicx}

\newcommand{\DEVELOPMENT}{0} 
\usepackage{ifthen}
\ifthenelse{\DEVELOPMENT = 1}{
	\newcommand{\mw}[1]{\textcolor{red}{\textbf{MW:} #1}}
}{
	\newcommand{\mw}[1]{}
}

\title{Measuring the Reliability of Hate Speech Annotations: \\ The Case of the European Refugee Crisis}

\author{Bj\"orn Ross {	} Michael Rist  {	} Guillermo Carbonell \\
 \bf Benjamin Cabrera {	}  Nils Kurowsky {	}  Michael Wojatzki\\
 \small Research Training Group ''User-Centred Social Media'' \\
 \small Department of Computer Science and Applied Cognitive Science \\
 \small University of Duisburg-Essen \\
 \tt firstname.lastname@uni-due.de
  }

\date{}

\begin{document}
\maketitle
\begin{abstract}
Some users of social media are spreading racist, sexist, and otherwise hateful content.
  For the purpose of training a hate speech detection system, the reliability of the annotations is crucial, but there is no universally agreed-upon definition.
  We collected potentially hateful messages and asked two groups of internet users to determine whether they were hate speech or not, whether they should be banned or not and to rate their degree of offensiveness. One of the groups was shown a definition prior to completing the survey.
  We aimed to assess whether hate speech can be annotated reliably, and the extent to which existing definitions are in accordance with subjective ratings.
  Our results indicate that showing users a definition caused them to partially align their own opinion with the definition but did not improve reliability, which was very low overall.
  We conclude that the presence of hate speech should perhaps not be considered a binary yes-or-no decision, and raters need more detailed instructions for the annotation.

\end{abstract}

\section{Introduction}
Social media are sometimes used to disseminate hateful messages.
In Europe, the current surge in hate speech has been linked to the ongoing refugee crisis.
Lawmakers and social media sites are increasingly aware of the problem and are developing approaches to deal with it, for example promising to remove illegal messages within 24 hours after they are reported \cite{titcomb2016}.

This raises the question of how hate speech can be detected automatically.
Such an automatic detection method could be used to scan the large amount of text generated on the internet for hateful content and report it to the relevant authorities.
It would also make it easier for researchers to examine the diffusion of hateful content through social media on a large scale.

From a natural language processing perspective, hate speech detection can be considered a classification task: given an utterance, determine whether or not it contains hate speech.
Training a classifier requires a large amount of data that is unambiguously hate speech.
This data is typically obtained by manually annotating a set of texts based on whether a certain element contains hate speech.

The reliability of the human annotations is essential, both to ensure that the algorithm can accurately learn the characteristics of hate speech, and as an upper bound on the expected performance \cite{warner2012,waseem2016hateful}.
As a preliminary step, six annotators rated 469 tweets. We found that agreement was very low (see Section 3).
We then carried out group discussions to find possible reasons. They revealed that there is considerable ambiguity in existing definitions.
A given statement may be considered hate speech or not depending on someone's cultural background and personal sensibilities.
The wording of the question may also play a role.
\mw{vll. sollten wir hier noch etwas mehr das Problem herausarbeiten: niedriges agreement, kein agreement, transfer, ambiguität}

We decided to investigate the issue of reliability further by conducting a more comprehensive study across a large number of annotators, which we present in this paper.

Our contribution in this paper is threefold:
\begin{itemize}
	\setlength\itemsep{0em}
	\item To the best of our knowledge, this paper presents the first attempt at compiling a German hate speech corpus for the refugee crisis.\footnote{Available at \url{https://github.com/UCSM-DUE/IWG_hatespeech_public}}
    \item We provide an estimate of the reliability of hate speech annotations.
    \item We investigate how the reliability of the annotations is affected by the exact question asked.
\end{itemize}

\section{Hate Speech}

For the purpose of building a classifier, \newcite{warner2012} define hate speech as ``abusive speech targeting specific group characteristics, such as ethnic origin, religion, gender, or sexual orientation''.
More recent approaches rely on lists of guidelines such as a tweet being hate speech if it ``uses a sexist or racial slur'' \cite{waseem2016hateful}.
These approaches are similar in that they leave plenty of room for personal interpretation, since there may be differences in what is considered offensive.
For instance, while the utterance \textit{``the refugees will live off our money''} is clearly generalising and maybe unfair, it is unclear if this is already hate speech.
More precise definitions from law are specific to certain jurisdictions and therefore do not capture all forms of offensive, hateful speech, see e.g. \newcite{matsuda1993}.
\label{twitterdef}
In practice, social media services are using their own definitions which have been subject to adjustments over the years \cite{jeong2016}.
As of June 2016, Twitter bans \emph{hateful conduct}\footnote{``You may not promote violence against or directly attack or threaten other people on the basis of race, ethnicity, national origin, sexual orientation, gender, gender identity, religious affiliation, age, disability, or disease.
We also do not allow accounts whose primary purpose is inciting harm towards others on the basis of these categories.'', The Twitter Rules}.

With the rise in popularity of social media, the presence of hate speech has grown on the internet.
Posting a tweet takes little more than a working internet connection but may be seen by users all over the world.

Along with the presence of hate speech, its real-life consequences are also growing.
It can be a precursor and incentive for hate crimes, and it can be so severe that it can even be a health issue \cite{burnap2014hate}.
It is also known that hate speech does not only mirror existing opinions in the reader but can also induce new negative feelings towards its targets \cite{martin2013}.
Hate speech has recently gained some interest as a research topic on the one hand -- e.g. \cite{nemanja2014,burnap2014hate,silva2016} -- but also as a problem to deal with in politics such as the \emph{No Hate Speech Movement} by the Council of Europe.

The current refugee crisis has made it evident that governments, organisations and the public share an interest in controlling hate speech in social media.
However, there seems to be little consensus on what hate speech actually is.

\section{Compiling A Hate Speech Corpus}
As previously mentioned, there is no German hate speech corpus available for our needs, especially not for the very recent topic of the refugee crisis in Europe.
We therefore had to compile our own corpus.
We used Twitter as a source as it offers recent comments on current events.
In our study we only considered the textual content of tweets that contain certain keywords, ignoring those that contain pictures or links.
This section provides a detailed description of the approach we used to select the tweets and subsequently annotate them.

To find a large amount of hate speech on the refugee crisis, we used 10 hashtags\footnote{\emph{\#Pack}, \emph{\#Aslyanten}, \emph{\#WehrDich}, \emph{\#Krimmigranten}, \emph{\#Rapefugees}, \emph{\#Islamfaschisten}, \emph{\#RefugeesNotWelcome}, \emph{\#Islamisierung}, \emph{\#AsylantenInvasion}, \emph{\#Scharia}} that can be used in an insulting or offensive way.
Using these hashtags we gathered 13\,766 tweets in total, roughly dating from February to March 2016.
However, these tweets contained a lot of non-textual content which we filtered out automatically by removing tweets consisting solely of links or images.
We also only considered original tweets, as retweets or replies to other tweets might only be clearly understandable when reading both tweets together.
In addition, we removed duplicates and near-duplicates by discarding tweets that had a normalised \textit{Levenshtein} edit distance smaller than .85 to an aforementioned tweet.
A first inspection of the remaining tweets indicated that not all search terms were equally suited for our needs.
The search term \emph{\#Pack} (vermin or lowlife) found a potentially large amount of hate speech not directly linked to the refugee crisis. It was therefore discarded.
As a last step, the remaining tweets were manually read to eliminate those which were difficult to understand or incomprehensible.
After these filtering steps, our corpus consists of 541 tweets, none of which are duplicates, contain links or pictures, or are retweets or replies.

As a first measurement of the frequency of hate speech in our corpus, we personally annotated them based on our previous expertise.
The 541 tweets were split into six parts and each part was annotated by two out of six annotators in order to determine if hate speech was present or not.
The annotators were rotated so that each pair of annotators only evaluated one part.
Additionally the offensiveness of a tweet was rated on a 6-point Likert scale, the same scale used later in the study.

Even among researchers familiar with the definitions outlined above, there was still a low level of agreement (Krippendorff's $\alpha =
.38$).
This supports our claim that a clearer definition is necessary in order to be able to train a reliable classifier.
The low reliability could of course be explained by varying personal attitudes or backgrounds, but clearly needs more consideration.

\section{Methods}
In order to assess the reliability of the hate speech definitions on social media more comprehensively, we developed two online surveys in a between-subjects design. They were completed by 56 participants in total (see Table \ref{tab:summary}).
The main goal was to examine the extent to which non-experts agree upon their understanding of hate speech given a diversity of social media content.
We used the Twitter definition of \textit{hateful conduct} in the first survey.
This definition was presented at the beginning, and again above every tweet.
The second survey did not contain any definition.
Participants were randomly assigned one of the two surveys.

The surveys consisted of 20 tweets presented in a random order. For each tweet, each participant was asked three questions.
Depending on the survey, participants were asked \textbf{(1)} to answer (yes/no) if they considered the tweet hate speech, either based on the definition or based on their personal opinion.
Afterwards they were asked \textbf{(2)} to answer (yes/no) if the tweet should be banned from Twitter.
Participants were finally asked \textbf{(3)} to answer how offensive they thought the tweet was on a 6-point Likert scale from 1 (Not offensive at all) to 6 (Very offensive). If they answered 4 or higher, the participants had the option to state which particular words they found offensive.

After the annotation of the 20 tweets, participants were asked to voluntarily answer an open question regarding the definition of hate speech.
In the survey with the definition, they were asked if the definition of Twitter was sufficient.
In the survey without the definition, the participants were asked to suggest a definition themselves.
Finally, sociodemographic data were collected, including age, gender and more specific information regarding the participant's political orientation, migration background, and personal position regarding the refugee situation in Europe.

The surveys were approved by the ethical committee of the Department of Computer Science and Applied Cognitive Science of the Faculty of Engineering at the University of Duisburg-Essen.

\section{Preliminary Results and Discussion}

Since the surveys were completed by 56 participants, they resulted in 1120 annotations.
Table \ref{tab:summary} shows some summary statistics.

\begin{table}[h]
  \centering
  \setlength{\tabcolsep}{0.4em}
  \begin{tabular}{lrrrr}
	\hline
	 & Def. & No def. & p  & r\\
    \hline
    Participants & 25   & 31 &   \\
    Age (mean)        & 33.3 & 30.5 & \\
    Gender (\% female) & 43.5 & 58.6 & \\
	\hline
	Hate Speech (\% yes) & 32.6     & 40.3 & .26 & .15 \\
	Ban  (\% yes)   & 32.6    & 17.6 & .01 & -.32 \\
	Offensive (mean) & 3.49       & 3.42 & .55 & -.08 \\
	\hline
	\end{tabular}
	  \caption{Summary statistics with p values and effect size estimates from WMW tests. Not all participants chose to report their age or gender.}
	  \label{tab:summary}
	\end{table}

To assess whether the definition had any effect, we calculated, for each participant, the percentage of tweets they considered hate speech or suggested to ban and their mean offensiveness rating. This allowed us to compare the two samples for each of the three questions. Preliminary Shapiro-Wilk tests indicated that some of the data were not normally distributed. We therefore used the Wilcoxon-Mann-Whitney (WMW) test to compare the three pairs of series. The results are reported in Table \ref{tab:summary}.

Participants who were shown the definition were more likely to suggest to ban the tweet.
In fact, participants in group one very rarely gave different answers to questions one and two (18 of 500 instances or 3.6\%).
This suggests that participants in that group aligned their own opinion with the definition.

We chose Krippendorff's $\alpha$ to assess reliability, a measure from content analysis, where human coders are required to be interchangeable. Therefore, it measures agreement instead of association, which leaves no room for the individual predilections of coders. It can be applied to any number of coders and to interval as well as nominal data. \cite{krippendorff2004}

\begin{figure}[h]
    \centering
    \includegraphics[width=0.5\textwidth]{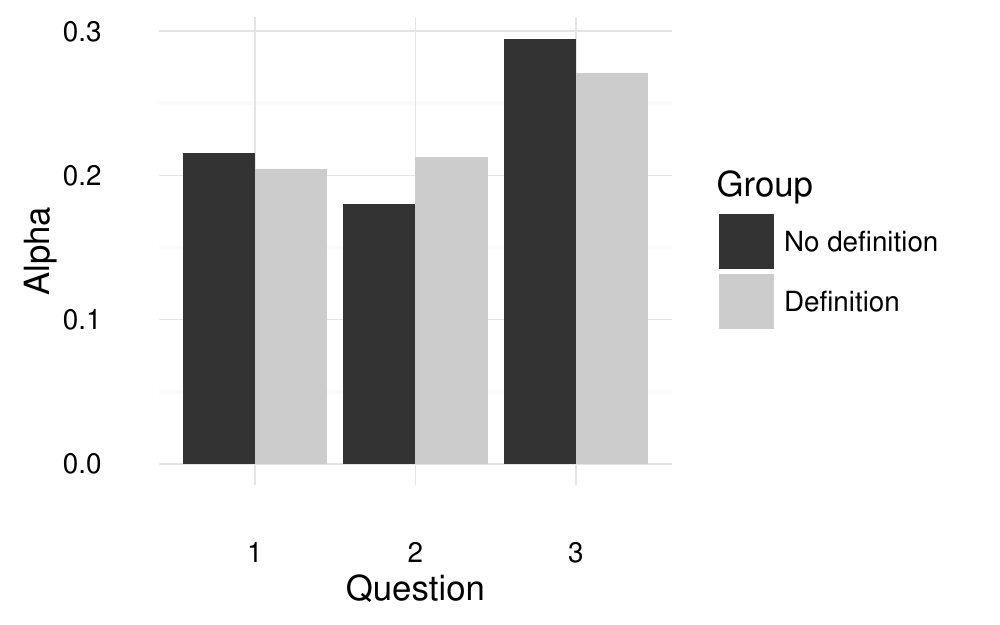}
    \caption{Reliability (Krippendorff's $\alpha$) for the different groups and questions}
    \label{fig:alphas}
\end{figure}
This allowed us to compare agreement between both groups for all three questions.
Figure \ref{fig:alphas} visualises the results.
Overall, agreement was very low, ranging from $\alpha = .18$ to $.29$.
In contrast, for the purpose of content analysis, Krippendorff recommends a minimum of  $\alpha = .80$, or a minimum of $.66$ for applications where some uncertainty is unproblematic \cite{krippendorff2004}.
Reliability did not consistently increase when participants were shown a definition.

To measure the extent to which the annotations using the Twitter definition (question one in group one) were in accordance with participants' opinions (question one in group two), we calculated, for each tweet, the percentage of participants in each group who considered it hate speech, and then calculated Pearson's correlation coefficient.
The two series correlate strongly ($r = .895, p < .0001$), indicating that they measure the same underlying construct.

\section{Conclusion and Future Work}

This paper describes the creation of our hate speech corpus and offers first insights into the low agreement among users when it comes to identifying hateful messages.
Our results imply that hate speech is a vague concept that requires significantly better definitions and guidelines in order to be annotated reliably.
Based on the present findings, we are planning to develop a new coding scheme which includes clear-cut criteria that let people distinguish hate speech from other content.

Researchers who are building a hate speech detection system might want to collect multiple labels for each tweet and average the results.
Of course this  approach does not make the original data any more reliable \cite{krippendorff2004}. Yet, collecting the opinions of more users gives a more detailed picture of objective (or intersubjective) hatefulness.
For the same reason, researchers might want to consider hate speech detection a regression problem, predicting, for example, the degree of hatefulness of a message, instead of a binary yes-or-no classification task.

In the future, finding the characteristics that make users consider content hateful will be useful for building a model that automatically detects hate speech and users who spread hateful content, and for determining what makes users disseminate hateful content.
\section*{Acknowledgments}
This work was supported by the Deutsche Forschungsgemeinschaft (DFG) under grant No. GRK 2167, Research Training Group ''User-Centred Social Media''.

\bibliographystyle{konvens2016}
\bibliography{hatespeech_bibliography.bib}

\end{document}